\documentclass[letterpaper,11pt]{article}
\usepackage[margin=1in]{geometry}
\usepackage[round]{natbib}
\usepackage{graphicx}
\graphicspath{{figs/}}
\usepackage{booktabs}
\usepackage{amsmath}
\usepackage{url}
\newcommand{\moep}{\texttt{moep}}
\newcommand{\secref}[2]{Section~\ref{#1}}
\newcommand{\figwidth}{0.62\textwidth}
\newcommand{\arxivonly}[1]{#1}
\usepackage{tikz}
\usetikzlibrary{arrows.meta,positioning}

\title{Half the Experts, All the Code:\\One-Shot Domain Pruning of
Mixture-of-Experts LLMs for Coding\thanks{Preprint. Under review. Code,
per-point expert selections, and quantized specialist checkpoints:
\texttt{github.com/anik-jha/moep}.}}
\author{Anik Jha$^{1}$\\[4pt]
  $^{1}$Independent Researcher\\
  \texttt{anik.k.jha@gmail.com}}
\date{}

\begin{document}
\maketitle

\begin{center}\textbf{Abstract}\end{center}
\vspace{-0.6em}
\begin{list}{}{\setlength{\leftmargin}{2.2em}\setlength{\rightmargin}{2.2em}}
\item\relax
The strongest open-weight coding models are mixture-of-experts (MoE)
networks: most of their size comes from large pools of alternative
``expert'' subnetworks, of which only a few act on any given token. That
pool is why these models do not fit on the machines most developers own,
yet for a user who only wants coding help, most experts encode abilities
that will never be invoked. We ask how many experts can be removed, and how
to choose which, by pruning two recent open-weight MoE models from
different families (Qwen3.6-35B-A3B and Gemma-4-26B-A4B) under five
selection strategies, and judging the result the way a user would: by
whether the model still writes correct code. Half the experts can be
removed from either model with no statistically detectable loss on the
primary code-generation benchmark, and the damage lands almost entirely on
abilities outside coding, which is the intended trade. But the selection
strategy that achieves this flips between the two models: the winner on
one trails significantly on the other, whose own winner collapses outright
at deeper pruning on the first, so a pruning recipe validated on one model
family cannot be assumed to work on another. We further show that
perplexity, the metric much of the pruning literature leans on, can rate a
broken model above an intact one; that a lightweight fine-tune recovers
about half of what aggressive pruning loses; and that against simply
quantizing the full model to the same memory budget, pruning wins only
where quantization would have to drop below 3 bits per weight. Five
attempts to overturn that crossover, with failure criteria fixed in
advance (better calibration, guarded selection, causally measured expert
importance, failure attribution, and an agentic evaluation that lets each
model repair its failures from execution feedback), all leave it
standing, and the last
shows single-shot benchmarks overstate compression penalties broadly:
one repair turn erases the 2-bit quantization penalty entirely. Expert
pruning works, but it demands per-model validation on the task the model
will actually serve.
\end{list}
\vspace{0.4em}

% Shared paper body: \input by main.tex (arXiv single-column) and aaai/main_aaai.tex
% (AAAI-27 two-column, anonymized). No preamble or abstract here.
\section{Introduction}

The best open-weight coding models are sparse mixtures of experts. The
pattern is consistent across model families: a modest number of active
parameters rides on a very large pool of experts, and that pool is what
determines whether the model fits in memory at all. A 35B-parameter MoE with
3B active parameters streams quickly enough on a laptop-class memory bus,
but only if its weights fit beside the KV cache in the first place. For the
current generation of open coding models, they usually do not.

\arxivonly{Two terms carry the paper. An MoE layer holds a pool of
parallel \emph{expert} subnetworks (128 or 256 in the models here) and a
small router that sends each token through eight of them, so most weights
sit idle on any given input; pruning here means deleting experts outright,
which shrinks the model on disk and in memory. We measure coding ability
with HumanEval+ and MBPP+, two standard suites of Python problems scored
by executing the generated code against tests (pass@1: the fraction solved
on the first attempt).} Our subjects are recent open-weight MoE releases
from two major model families with substantially different internal
designs; agreement across them is the minimum bar any portable pruning
recipe must clear, and, as we will show, current recipes do not clear it.

This is a peculiarly wasteful failure mode. The expert pool is large
because the model is general: a developer who wants a coding assistant
pays the memory cost of experts serving multilingual chat, creative
writing, and factual recall they will rarely invoke. Recent work
has shown that for a narrow enough task the redundancy is enormous: half of
all experts can be removed with negligible loss for machine translation
\citep{martin2026translation}, and one-shot criteria such as REAP retain
almost all coding ability at 50\% compression on coding-oriented models
\citep{lasby2026reap}.

We study this line of work in the direction that matters in practice:
extracting coding specialists, evaluating them the way coding models are
used (code generation, greedy decoding, quantized weights), and sizing
them for machines people own. Along the way we ask which selection signal
carries the result. A natural idea, which we introduce and test, is a
\emph{domain-contrastive} score: rank each expert by the \emph{ratio} of
its target-domain to general-domain importance, separating experts that
matter for coding from experts that matter for everything. It does
separate them, but the experiment shows the separation is the wrong thing
to keep.

We build a pipeline (\moep) that profiles routing and expert activations,
ranks experts under interchangeable criteria, performs checkpoint surgery
without loading the model, heals aggressive prunes by distillation, and
exports quantized GGUF artifacts verified on real hardware. With it we
make four contributions:

\begin{itemize}
\item A two-family study of coding-specialist extraction, on
Qwen3.6-35B-A3B (a hybrid-attention multimodal MoE whose expert-pruning
behavior had previously been examined only in the biomedical domain
\citep{biomed2026reliability}) and Gemma-4-26B-A4B, with identical
calibration, criteria, surgery invariants, and functional evaluation on
both, and degeneration-rate tracking alongside utility metrics.
\item Evidence that one-shot criterion rankings do not transfer across
model families: the criterion that is best on one family is significantly
worse than a competitor on the other, in both directions (REAP beats
routing mass by 2.4 points on Qwen where routing mass later collapses;
routing mass beats REAP by 12.8 points, CI $[+6.7, +18.9]$, on Gemma). The
practical consequence is that a deployed specialist requires per-model
criterion validation with functional evals; no published one-shot criterion
can be assumed to carry over.
\item A design principle that \emph{does} transfer, from a controlled
comparison including a domain-contrastive score we introduce as a
candidate: expert \emph{magnitude}-family signals are load-bearing, and
domain \emph{informativeness} can complement but not replace them (random
selection collapses on both families; a dose-response sweep shows quality
falling as contrast displaces magnitude). Alongside it, three independent
demonstrations that perplexity misleads pruning studies: better-than-base
perplexity with a 58-point pass@1 deficit, rank inversions within a
criterion family, and a chat-template artifact that inflates measured
perplexity up to 30-fold per sample (14x over the full holdout).
\item A matched comparison of pruning against quantization at equal memory
with paired confidence intervals, which locates a crossover rather than a
winner: quantizing the full model is never worse and is significantly
better on MBPP+ wherever a 3-bit-or-above quantization fits the budget;
the pruned specialist wins only below that boundary (at $\sim$13~GB it
significantly beats the 2-bit full model on HumanEval+). Sizes and
throughput are measured on a 119~GB unified-memory device.
\end{itemize}

\section{Related Work}

\paragraph{One-shot expert pruning.} Structured pruning of MoE experts has
converged on cheap saliency statistics collected over a calibration set.
REAP scores each expert by router-gate-weighted activation norms and shows
that pruning dominates expert merging on generative tasks
\citep{lasby2026reap}. A recent unified formulation
\citep{unified2026scoring} organizes criteria as products of routing
frequency, gate mass, and activation norm, and finds that no single point in
this family wins everywhere: task-specific pruning benefits from retaining
frequency and gate signals, while task-agnostic pruning favors gate-free
norms. MAESTRO models cross-layer routing trajectories as a Markov chain
\citep{maestro2026}, and other work explores trajectory- and
calibration-free variants \citep{pathfinder2025,aimer2026}. All of these
rank by importance measured on a single distribution; none contrasts
importance across distributions, which is the gap we target.

\paragraph{Domain-specific specialists.} Extracting a specialist by pruning
experts is an active line. \citet{martin2026translation} extract
machine-translation specialists with routing-mass ranking and dynamic
per-layer capacity, pruning 50\% of experts with negligible loss and 75\%
after a short full fine-tune, and report that parameter-efficient recovery is
insufficient; our coding result is the opposite on that point.
\citet{yao2026cprune} cluster and merge experts within layers for domain
deployment, and \citet{dong2025easyep} identify domain-relevant experts from
a few in-domain demonstrations, keeping half the experts of large MoEs at
comparable in-domain performance. Closest in setup,
\citet{biomed2026reliability} evaluate six saliency criteria (including
REAP, EASY-EP, and random selection) across compression ratios on four MoE
models, one of which is the same Qwen3.6-35B-A3B we study, in the biomedical
domain, and show that factual reliability can degrade before utility metrics
move, which motivates our degeneration tracking. Our criterion comparison
differs from theirs in domain (coding), in evaluating functional code
generation under greedy decoding rather than biomedical generation and
classification utility, and in the contrastive criterion and its
dose-response analysis; on the shared model and criteria the two studies are
complementary rather than overlapping. Most directly relevant, \citet{he2026lessismoe} argue
that expert-\emph{level} pruning collapses on generative benchmarks including
HumanEval and MBPP, and propose finer-grained intermediate-dimension pruning
instead. Our results refine that finding: expert-level pruning does collapse when a
general model is pruned and evaluated across all domains, but with
calibration and evaluation matched to the target domain it is
near-lossless, which is exactly the regime a deployed specialist
occupies.

\paragraph{Serving compressed MoEs.} On consumer hardware the dominant
footprint lever is weight quantization, via the GGUF ecosystem
\citep{llamacpp} with importance-matrix calibration protecting the
most-used weights at low bit widths. Pruning and quantization are usually
treated as orthogonal and composed; we instead put them in direct
competition at fixed memory, both sides imatrix-calibrated, and find a
genuine tradeoff rather than a foregone conclusion in pruning's favor.

\section{Method}

\arxivonly{%
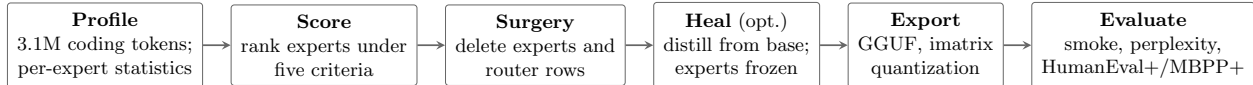
\begin{figure}[t]
\centering
\resizebox{\textwidth}{!}{%
\begin{tikzpicture}[
  stage/.style={draw=black!60, rounded corners=1.5pt, align=center,
                font=\footnotesize, minimum height=11mm, inner sep=1.6mm},
  arr/.style={-stealth, black!60, thick}]
\node[stage] (prof) {\textbf{Profile}\\ 3.1M coding tokens;\\ per-expert statistics};
\node[stage, right=4.5mm of prof] (score) {\textbf{Score}\\ rank experts under\\ five criteria};
\node[stage, right=4.5mm of score] (surg) {\textbf{Surgery}\\ delete experts and\\ router rows};
\node[stage, right=4.5mm of surg] (heal) {\textbf{Heal} (opt.)\\ distill from base;\\ experts frozen};
\node[stage, right=4.5mm of heal] (exp) {\textbf{Export}\\ GGUF, imatrix\\ quantization};
\node[stage, right=4.5mm of exp] (ev) {\textbf{Evaluate}\\ smoke, perplexity,\\ HumanEval+/MBPP+};
\draw[arr] (prof) -- (score);
\draw[arr] (score) -- (surg);
\draw[arr] (surg) -- (heal);
\draw[arr] (heal) -- (exp);
\draw[arr] (exp) -- (ev);
\end{tikzpicture}}
\caption{The pipeline. A calibration corpus is routed through the base model
once to collect per-expert statistics; experts are ranked and deleted
directly in the checkpoint (surviving weights verified byte-identical);
aggressive prunes are optionally healed by distilling from the base model
with the surviving experts frozen; artifacts are quantized and every
operating point is evaluated end to end on the target device.}
\label{fig:pipeline}
\end{figure}}

\subsection{Setup and profiling}

Consider an MoE layer $\ell$ with experts $\{1,\dots,E\}$ and a router that,
for token $x$, produces probabilities $p_\ell(x) \in \Delta^{E}$ (softmax
over all experts, in float32 in the models we study), selects the top-$k$,
and renormalizes their weights to obtain gates $g_{\ell,e}(x)$. Let
$f_{\ell,e}(x)$ denote the expert's unweighted output. Streaming a
calibration corpus $D$ through the model, we accumulate per (layer, expert),
in float32: the routed-token count $n_{\ell,e}$; the gate mass $\sum
g_{\ell,e}(x)$; the full-softmax probability mass $\sum p_{\ell,e}(x)$ over
\emph{all} tokens; and first and second moments of $\lVert
f_{\ell,e}(x)\rVert_2$ together with the REAP summand $g_{\ell,e}(x)\lVert
f_{\ell,e}(x)\rVert_2$. One pass therefore yields every criterion we
compare. Because grouped-expert implementations sum weighted expert
outputs in place, unweighted norms are not observable from module outputs;
we recompute them from the expert weights inside a forward hook, a
one-time profiling cost.

\subsection{Domain-contrastive scoring}

Define the expert's \emph{flow} on corpus $D$ as
\begin{equation}
F_{\ell,e}(D) \;=\; \tfrac{1}{|D|}\textstyle\sum_{x \in D_{\ell,e}}
g_{\ell,e}(x)\,\lVert f_{\ell,e}(x)\rVert_2 ,
\end{equation}
the per-token expected gated contribution of expert $e$ (the
frequency-weighted REAP statistic). Given a target corpus $T$ (agentic
coding) and a reference corpus $R$ (general chat and reasoning), we score
\begin{equation}
S^{\mathrm{con}}_{\ell,e} \;=\;
\frac{F_{\ell,e}(T)}{F_{\ell,e}(R) + \varepsilon},
\end{equation}
and keep the top-$K$ experts per layer. High $S^{\mathrm{con}}$ marks
domain specialists; experts with large but undifferentiated flow (generic
infrastructure the model needs for any input) score near 1 and survive only
if capacity permits, while experts important solely off-domain are removed
first. The reference statistics reuse the same profiling machinery, so the
extra cost over standard REAP is one profiling pass on $R$.

Both flows are measured under the base model's routing. After pruning, the
router renormalizes over survivors, so measured flows are estimates of
post-surgery behavior; the same caveat applies to every one-shot criterion,
and our equivalence tests (below) bound the part of the gap that is
mechanical rather than behavioral.

\subsection{Surgery, healing, and export}

The models we target store experts as fused three-dimensional parameters
(e.g.\ gate/up projections of shape $[E, 2d_{\mathrm{ff}}, d]$), so pruning
is an index-select on the leading dimension of two tensors per layer plus a
row slice of the router matrix, applied shard by shard without instantiating
the model; peak memory is one tensor. The pipeline verifies surviving rows
byte-for-byte during the copy. Two invariants, enforced in tests against a
synthetic model of the same class, guard correctness: pruning zero experts
must reproduce base logits exactly, and a pruned model must produce logits
identical to the base model with dropped experts masked out of its router,
which holds because restricting a softmax to survivors and renormalizing
over the selected top-$k$ commute.

Exported specialists are converted to GGUF and quantized; every artifact is
load-verified by serving one completion and benchmarked on the target
device before being reported.

\section{Experimental Setup}

\paragraph{Models.} Qwen3.6-35B-A3B \citep{qwen36} is our primary subject:
35B total and roughly 3B active parameters, 40 layers, 256 experts per
layer with top-8 routing plus one always-on shared expert, hidden size
2048, per-expert FFN width 512, a 3:1 interleave of linear attention
(gated deltanet) and full attention, a vision tower, and a multi-token
prediction head carrying its own expert copies (ignored at load by the
reference implementation; we drop it from pruned exports and note this in
the artifact card). Gemma-4-26B-A4B \citep{gemma4} is the replication
family: 30 layers, 128 experts per layer with top-8 routing, hidden size
2816, per-expert FFN width 704, with a dense per-layer MLP combined
additively with the MoE output serving the always-on role, and a router
that applies a learned per-expert scale after top-$k$ renormalization;
surgery therefore slices that scale vector alongside the router rows, and
because the router reads a different tensor than the experts, its
full-softmax probabilities are captured from the router module directly
during profiling. All architecture facts were read from the released
checkpoints' configuration, tensor index, and installed modeling source
rather than from documentation, which we found necessary in practice:
secondary sources disagreed on Qwen's expert count by a factor of two.

\arxivonly{%
\begin{table}[t]
\centering
\small
\begin{tabular}{lll}
\toprule
 & Qwen3.6-35B-A3B & Gemma-4-26B-A4B \\
\midrule
Total / active params & 36.0B / $\approx$3B & 25.8B / $\approx$4B \\
MoE layers $\times$ experts & 40 $\times$ 256 & 30 $\times$ 128 \\
Routing & top-8 + shared expert & top-8 + parallel dense MLP \\
Attention & 3:1 linear\,:\,full & sliding + periodic full \\
\midrule
\multicolumn{3}{l}{Benchmarks: HumanEval+ (164 problems) and MBPP+ (378), greedy pass@1} \\
\multicolumn{3}{l}{Calibration: 2{,}047 chat samples, 70/30 coding-to-general, 3.15M tokens} \\
\multicolumn{3}{l}{Hardware: one 119\,GB unified-memory device; all artifacts served by llama.cpp} \\
\bottomrule
\end{tabular}
\caption{The study at a glance. Active-parameter counts follow the models'
A3B/A4B designations; all other facts are read from the released
checkpoints.}
\label{tab:setup}
\end{table}}

\paragraph{Calibration.} The target mix $T$ holds 2{,}047 chat-formatted
samples: 717 from Evol-CodeAlpaca, 358 tool-calling dialogues (Hermes
function calling; the originally planned xLAM set is gated), 358 SWE-agent
trajectories from SWE-smith, and 614 general samples from Tulu-3 SFT as
ballast (a 70/30 split of code to general), packed at 4{,}096 tokens without
padding into 3.15M profiled tokens. The reference corpus $R$ is the
general slice, profiled separately.
Held-out sets of 256 coding and 256 general samples support perplexity
evaluation; they never appear in calibration. Every expert of every layer
received traffic during profiling on Qwen (minimum layer-expert coverage
100\%); on Gemma the minimum is 99.9\%, i.e.\ a handful of experts among
3{,}840 saw no calibration token. Both are consistent with the auxiliary
load-balancing losses reported for these model families.

\paragraph{Criteria and ratios.} We compare contrastive scoring against
REAP (mean gated norm), routing mass (equivalently gate mass, the criterion
of \citet{martin2026translation}), mean activation norm (MAN), and seeded
random keeps, at keep ratios 0.75, 0.5, 0.375, and 0.25, uniform per layer.
Uniform allocation keeps exports compatible with standard single
expert-count metadata in current serving stacks; per-layer capacity
allocation is left to future work.

\paragraph{Evaluation.} Stage one is a fixed 8-prompt greedy smoke test
with a repetition/emptiness detector; its degeneration count is recorded
and reported but does not gate scoring, because we found it fails in both
directions (a checkpoint with a clean 8/8 smoke can score 0.09 pass@1,
and a checkpoint failing one prompt can be perfectly usable). Stage two is
packed perplexity on the code and general holdouts (48 sequences of
2{,}048 tokens each); for Gemma we measure it on bare message content,
because its chat template inflates teacher-forced perplexity up to
roughly 30-fold on individual samples and 14x over the full holdout
(\secref{sec:ppl-fragility}{perplexity-fragility}). Stage three
is HumanEval+ and MBPP+ \citep{evalplus} pass@1, greedy, with reasoning
mode disabled for base and pruned models alike, generated against
identically quantized (Q8\_0) weights served by the same llama.cpp build,
so quantization and template handling cancel in comparisons. Stage four
measures memory and throughput of deployable Q4\_K\_M artifacts. Because
all generation is greedy, repeated runs of one checkpoint are identical;
the uncertainty we report is therefore the 95\% paired bootstrap
confidence interval over benchmark items (about $\pm 3$ to $\pm 4$ points
on the 164-item HumanEval+ at these scores), given alongside every
comparison, plus mean and standard deviation across selection seeds where
the criterion itself is stochastic (random). We
additionally report the degeneration rate (empty or looping generations)
at every point, following the reliability concerns of
\citet{biomed2026reliability}. All hardware numbers come from a single
NVIDIA DGX Spark (GB10, 119~GB unified memory at approximately 273~GB/s),
chosen deliberately: it is representative of the unified-memory
consumer-class devices this work targets.

\section{Results}

\subsection{The specialist trade made explicit}

Removing 128 of 256 experts per layer with coding-heavy REAP scoring
produces a 19.0B-parameter model whose code perplexity rises from 5.378 to
5.492, an increase of 2.1\%, while perplexity on held-out general chat
rises from 5.201 to 12.004, an increase of 131\%. On HumanEval+ the
specialist scores 0.896 against the base model's 0.890. Under domain-focused
calibration, then, half the expert pool was carrying capabilities the coding
distribution never exercises, and the damage from removing it concentrates
almost entirely off-domain.

\begin{table}[t]
\centering
\small
\setlength{\tabcolsep}{4pt}
\begin{tabular}{lrrrrr}
\toprule
Keep & Params & \multicolumn{2}{c}{Ppl (code/gen)} & HE+ & MBPP+ \\
\midrule
base   & 36.0B & 5.38 & 5.20  & 0.890 & 0.772 \\
75\%   & 27.0B & 5.42 & 5.80  & 0.890 & 0.759 \\
50\%   & 19.0B & 5.49 & 12.00 & 0.896 & 0.728 \\
37.5\% & 15.0B & 5.89 & 30.99 & 0.860 & 0.733 \\
25\%   & 10.9B & 5.45 & 30.92 & 0.787 & 0.659 \\
\bottomrule
\end{tabular}
\caption{REAP frontier on Qwen3.6-35B-A3B with coding-heavy calibration,
greedy pass@1 at Q8\_0 on both HumanEval+ and MBPP+. HumanEval+ is
statistically unchanged through 50\% keep and marginally degraded at 37.5\%
(paired CIs in the text; pruned models exceeding base by 1 to 2 points is the
noise floor of a 164-item benchmark, not a claim). MBPP+ is unchanged at
75\% keep ($-1.3$, CI $[-3.7, +1.1]$) but pays a significant cost from
50\% keep down ($-4.5$, CI $[-7.7, -1.6]$, $p{=}0.006$). General-chat
perplexity degrades 2.3x at 50\% keep and 6x below it. No point produced
degenerate smoke generations.}
\label{tab:frontier}
\end{table}

Table~\ref{tab:frontier} extends this to the full frontier
(Figure~\ref{fig:frontier-reversal}a plots it beside the second family's). Two things
stand out. First, the flat region is wide, and we can be precise about
where it ends. At 75\% and 50\% keep the specialist is statistically
indistinguishable from base: paired per-task comparisons show 8 and 9 of
164 HumanEval+ outcomes flipping (in both directions), with bootstrap
differences of $0.0$ and $+0.6$ points and 95\% CIs of $[-3.1, +3.7]$
and $[-3.1, +4.3]$ respectively. We stress that ``unchanged'' means
exactly that, not improvement. At 37.5\% keep the first cost appears: a
3.0-point HumanEval+ drop (9 flips against, 4 for; CI $[-7.3, +1.2]$)
that does not reach significance on 164 items but is directionally
consistent. The sharp break comes only at 25\% keep. Domain-focused
calibration buys real headroom, but only up to a point. MBPP+ traces the
same frontier shape one notch earlier: unchanged at 75\% keep, but already
paying a significant 4.5 points at 50\% keep ($p{=}0.006$; 9 flips for, 26
against on 378 items). The two benchmarks therefore disagree about
whether the 50\% point is free on this family, and we report both rather
than let HumanEval+ speak alone; on Gemma the corresponding best-criterion
50\% point is statistically unchanged on both benchmarks
(\secref{sec:gemma}{cross-family}).
Second, perplexity and functional ability decouple at the
aggressive end: the 25\%-keep model has \emph{lower} code perplexity than
the 37.5\% one (5.45 vs.\ 5.89) yet loses seven points of pass@1. Past the flat region, next-token fit on code stops predicting whether
the model can still complete a function correctly. General perplexity
tells the other half of the story: the specialist sheds general ability
early (2.3x at 50\%) and severely (6x) beyond that, which is the
intended trade, not a defect.

\subsection{Criterion comparison}

Before any downstream evaluation, the criteria already reveal structure.
At 50\% keep, the selections of REAP and MAN overlap by 92.0\% of kept
experts, and routing frequency and gate mass overlap by 94.4\%, but the
two families agree on only 56 to 64\% of their keeps. The norm family and
the frequency family are measuring genuinely different things, and at
least 36\% of the model's experts live or die by the choice between them.

\begin{table}[t]
\centering
\small
\begin{tabular}{lcccc}
\toprule
& \multicolumn{2}{c}{keep 50\%} & \multicolumn{2}{c}{keep 25\%} \\
Criterion & HE & HE+ & HE & HE+ \\
\midrule
REAP           & \textbf{0.939} & \textbf{0.896} & \textbf{0.835} & \textbf{0.787} \\
MAN            & 0.915 & 0.884 & 0.762 & 0.720 \\
Routing mass   & 0.896 & 0.872 & \multicolumn{2}{c}{collapse (2/8)} \\
Contrastive    & 0.207 & 0.201 & \multicolumn{2}{c}{collapse (6/8)} \\
Random (seeded)& 0.360 & 0.311 & \multicolumn{2}{c}{collapse (5/8)} \\
\midrule
base           & 0.915 & 0.890 & 0.915 & 0.890 \\
\bottomrule
\end{tabular}
\caption{Criterion shootout on Qwen3.6-35B-A3B, greedy pass@1 at Q8\_0.
``Collapse ($k$/8)'' marks models with $k$ of 8 fixed smoke prompts
degenerate (empty or looping) that additionally could not complete code
evaluation (the 25\%-keep random model cannot produce a well-formed
completion for a one-line function). Smoke degeneration is reported, not
used as a gate: the 50\%-keep random model passes 7 of 8 smoke prompts
and still scores 0.311.}
\label{tab:shootout}
\end{table}

Table~\ref{tab:shootout} is the downstream verdict, and it reorders the
intuitions above. The gated-norm criterion (REAP) wins at both ratios and
its margin grows exactly where selection gets hard: at 25\% keep it holds
a 6.7-point advantage over MAN, its 92\%-overlap sibling, meaning the
last sliver of disagreement between the two is where the damage lives.
Routing mass, the winning criterion for translation specialists in prior
work, is competitive at 50\% but collapses outright at 25\% (on this
family; \secref{sec:gemma}{cross-family} shows the ranking inverts on another).
Random selection is the load-bearing baseline, so we score it rather than
disqualify it: at 50\% keep it reaches only 0.311 HumanEval+ ($-57.9$
points versus base, CI $[-65.8, -50.0]$) and 0.278 MBPP+ ($-49.5$,
CI $[-55.0, -43.9]$), and at 25\% keep it cannot complete a one-line
function. The collapse is not a property of one draw: across three
selection seeds random-50 scores $0.268 \pm 0.046$ on Qwen and
$0.142 \pm 0.121$ on Gemma (mean $\pm$ standard deviation), with no seed
approaching any informed criterion. At these compression levels there is
no free lunch from MoE redundancy alone; the selection criterion carries
the result.
We return to this model in
\secref{sec:ppl-fragility}{perplexity-fragility}, because by held-out
perplexity it looks \emph{better} than base.

The contrastive criterion we introduce is the one worth dwelling on. Used alone it destroys the model at both ratios (its 50\%-keep
survivor of the smoke gate emits repetition loops on real tasks and scores
0.207). The mechanism is visible in its selections: by scoring pure domain
informativeness, the ratio demotes exactly the high-flow, domain-neutral
experts (score near 1) that carry shared computation for every input, syntax
and formatting included. Removing informative-but-small experts costs
little; removing uninformative-but-large ones is fatal.

We tested whether a milder dose of the same signal helps, rather than
assuming it. A magnitude-gated hybrid that uses contrast only to break ties
among experts of comparable flow ($S = F_T^2/(F_T+F_R)$) recovers much of
the lost ground but still trails REAP substantially at 25\% keep (0.470
versus 0.787 HumanEval+), and a rank-blend variant fails the smoke gate
outright. The three points trace a monotonic dose-response: as domain
contrast displaces magnitude in the ranking, quality falls
(REAP $0.787 \rightarrow$ magnitude-gated hybrid $0.470 \rightarrow$ pure
contrast, collapse). We read this as a design principle for one-shot
selection rather than a failed criterion: expert \emph{magnitude} is the
load-bearing signal, and domain \emph{informativeness} can reweight it at
the margins but cannot substitute for it. This sharpens, on the generative
coding task, the selection principle of \citet{unified2026scoring} that
task-specific pruning must retain magnitude-carrying signals.

\subsection{Cross-family replication: the ranking does not transfer}
\label{sec:gemma}

\begin{figure}[t]
\centering
\includegraphics[width=\figwidth]{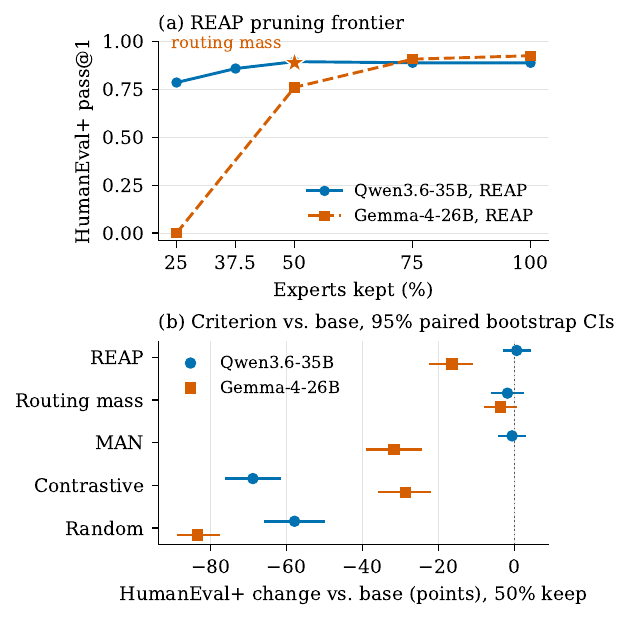}
\caption{(a) HumanEval+ against experts kept under REAP (greedy pass@1 at
Q8\_0); the star is Gemma's routing-mass selection at 50\% keep.
(b) Criterion-minus-base differences at 50\% keep, 95\% paired-bootstrap
CIs: REAP matches base on Qwen and trails by 16.5 on Gemma, where routing
mass is near base; random collapses on both.}
\label{fig:frontier-reversal}
\end{figure}

Every claim so far is one family's. We repeated the study end to end on
Gemma-4-26B-A4B: same calibration mix, same profiling statistics, same
surgery invariants (enforced against a tied-embedding synthetic model of
the Gemma class), same greedy Q8\_0 evaluation.
Table~\ref{tab:gemma} and Figure~\ref{fig:frontier-reversal} show the
result.

\begin{table}[t]
\centering
\small
\setlength{\tabcolsep}{3pt}
\begin{tabular}{lrrrrr}
\toprule
Point & Params & \multicolumn{2}{c}{Ppl (code/gen)} & HE+ & MBPP+ \\
\midrule
base            & 25.8B & 12.2 & 26.3  & 0.927 & 0.749 \\
REAP 75\%       & 20.1B & 10.6 & 18.5  & 0.909 & 0.741 \\
Routing mass 50\% & 14.4B & 18.5 & 91.9 & \textbf{0.890} & \textbf{0.741} \\
REAP 50\%       & 14.4B & 25.6 & 55.8  & 0.762 & 0.511 \\
Contrastive 50\% & 14.4B & 22.3 & 128.2 & 0.640 & 0.534 \\
MAN 50\%        & 14.4B & 24.5 & 118.2 & 0.610 & 0.577 \\
Random 50\%     & 14.4B & 33.9 & 54.6  & 0.091 & 0.254 \\
REAP 25\%       & 8.7B  & 47.0 & 123.5 & 0.000 & 0.005 \\
\bottomrule
\end{tabular}
\caption{Gemma-4-26B-A4B, greedy pass@1 at Q8\_0, coding-heavy calibration.
Perplexity is measured on bare message content
(\secref{sec:ppl-fragility}{perplexity-fragility}). Bold marks the best pruned point at
50\% keep: routing mass is statistically indistinguishable from base on
both benchmarks (HE+ $-3.7$, CI $[-7.9, +0.6]$, $p{=}0.15$; MBPP+ $-0.8$,
CI $[-3.7, +2.4]$, $p{=}0.74$).}
\label{tab:gemma}
\end{table}

The headline result replicates: removing half the experts of a general MoE
leaves coding functionally unchanged, on a second architecture, under the
same calibration recipe. So does the specialist trade (general perplexity
rises 3.5-fold at the best 50\% point against 1.5-fold on code), the
random-selection collapse ($0.142 \pm 0.121$ HumanEval+ across three
selection seeds at 50\% keep, zero at 25\%),
and the failure of contrastive scoring to match magnitude-family criteria.

What does not replicate is \emph{which criterion wins}, and the reversal
is not subtle. REAP, dominant on Qwen at every ratio, drops 16.5 points at
50\% keep on Gemma (CI $[-22.6, -11.0]$) and collapses entirely at 25\%
keep, the ratio it survives best on Qwen. Routing mass, which collapses at
25\% keep on Qwen, is Gemma's best criterion, beating REAP by 12.8 points
(CI $[+6.7, +18.9]$, exact McNemar $p{=}10^{-4}$) and statistically
matching the base model on both benchmarks. The two families disagree
about the two leading published criteria in both directions. Selection
structure differs accordingly: on Gemma the REAP and routing-mass keep
sets overlap by only 52\% of kept experts, so the choice between them
decides the fate of nearly half the expert pool. We do not offer a mechanism; the architectures differ in expert count,
expert width, router scaling, and the dense pathway's role, and any of
these could move which statistic tracks importance. The operational
conclusion does not depend on one: one-shot criterion rankings, including
ours, do not transfer across families, and a deployed specialist requires
a per-model criterion shootout with functional evaluation. This is consistent with, and sharpens, the position of
\citet{unified2026scoring} that no single point in the criterion family
wins everywhere: we show the instability across model families at fixed
task and fixed domain, where prior work varied the task.

\subsection{Perplexity is not a viable gate for pruning studies}
\label{sec:ppl-fragility}

Three independent observations from this study argue that held-out
perplexity, the primary metric of much of the pruning literature, cannot
carry a pruning result on its own.

First, it inverts quality orderings. The random-pruned Qwen model of
Table~\ref{tab:shootout} scores \emph{better} held-out code perplexity
than the base model (4.82 versus 5.38 under the study's standard protocol)
while losing 57.9 points of pass@1: an unbiased random sample of half the
expert population preserves the output distribution that perplexity
measures, teacher forcing hides the compounding routing errors, and
free-running generation collapses into repetition. Within the REAP
frontier the same decoupling appears in milder form (the 25\%-keep model
has lower code perplexity than the 37.5\%-keep model and seven points less
pass@1), and on Gemma the templated-perplexity ranking across criteria
anti-correlates with the pass@1 ranking.

Second, it is protocol-fragile in ways that dwarf the effects under study.
Gemma's chat template inflates teacher-forced perplexity up to roughly
30-fold on individual samples (417 templated versus 13 plain on one
held-out sample) and 14x over the full 48-sequence holdout (170.8 versus
12.2), while the same base model generates HumanEval+ at 0.927; a
protocol that is routine on one family is meaningless on another. The
reverse choice misleads too: on bare content, instruction-tuned Qwen's
base model scores \emph{worse} than several of its own prunes, because
template-free chat text is itself off-distribution for the base model.
Both protocols are internally consistent within a family; neither
supports comparisons across families, and we report each family under the
protocol that its template behavior admits.

Third, cheap degeneracy gates are unreliable in both directions: a Gemma
prune with a clean 8/8 smoke test scores 0.091, and a Qwen prune failing
one smoke prompt of eight scores 0.311. In this study, functional
evaluation is the only measurement that consistently ranks checkpoints
the way a user would.

\subsection{Deployment: pruning against quantization at equal memory}
\label{sec:deploy}

Pruning is one way to shrink a model; quantizing the weights is another, and
a practitioner with a fixed memory budget has to choose. We put the two on
equal footing across the full budget range where the choice exists: the
Qwen specialist pruned to 50\% and quantized at several precisions against
the \emph{full} base model quantized to matching sizes, taking the base
quantizations from a public imatrix-calibrated release so the baseline is
the strongest low-bit version available; our own pruned checkpoints are
quantized with a matching imatrix so neither side is handicapped.
Table~\ref{tab:iso} reports the result with paired confidence intervals.

\begin{table}[t]
\centering
\small
\setlength{\tabcolsep}{4pt}
\begin{tabular}{llcc}
\toprule
Budget & Model & HE+ & MBPP+ \\
\midrule
$\sim$11--13.5 GB & pruned 50\%, Q4 (11.4 GB)   & 0.902 & 0.720 \\
                 & pruned 50\%, Q5 (13.3 GB)   & 0.915 & 0.749 \\
                 & base, IQ2\_M (13.0 GB)      & 0.896 & 0.730 \\
                 & base, Q2\_K (13.5 GB)       & 0.848 & 0.754 \\
\midrule
$\sim$17--22 GB   & base, Q3\_K\_M (17.1 GB)    & 0.890 & \textbf{0.775} \\
                 & pruned 50\%, Q8 (19.75 GB)  & 0.896 & 0.728 \\
                 & base, Q4\_K\_M (22.3 GB)    & 0.884 & 0.759 \\
\bottomrule
\end{tabular}
\caption{Pruning versus quantization at matched memory (Qwen3.6-35B-A3B),
greedy pass@1, all quantizations imatrix-calibrated. Only two comparisons
in this table are statistically significant: the pruned Q5 beats the
2-bit base on HumanEval+ ($+6.7$, CI $[+3.0, +11.0]$, $p{=}0.003$), and
the 3-bit base beats the pruned Q8 on MBPP+ ($-4.8$ for pruning, CI
$[-7.7, -2.1]$, $p{=}0.0009$; bolded). Every other pairing, including
every HumanEval+ comparison in the upper tier against IQ2\_M, is within
noise on these benchmarks.}
\label{tab:iso}
\end{table}

\arxivonly{%
\begin{figure}[t]
\centering
\includegraphics[width=0.95\textwidth]{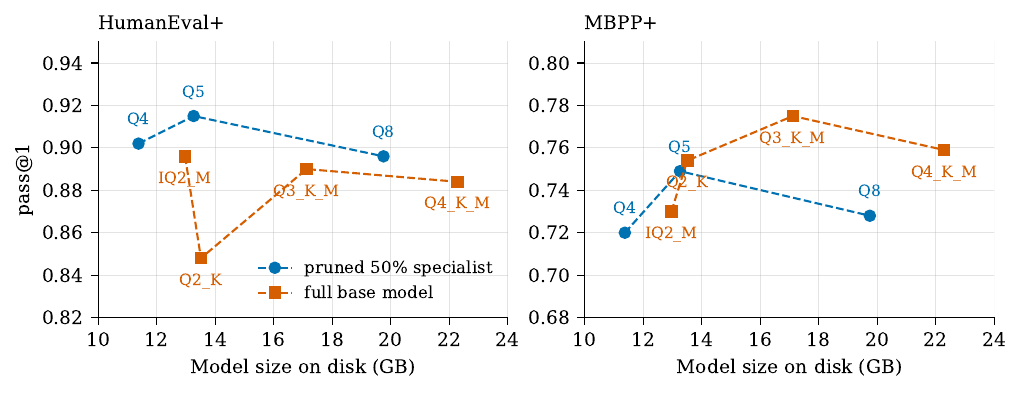}
\caption{Accuracy against model size for every deployable artifact in
Table~\ref{tab:iso}: the 50\%-pruned specialist at 4/5/8-bit quantization
(circles) versus the full base model at 2/3/4-bit (squares). Left of
$\sim$17\,GB, base quantizations must drop below 3 bits and the pruned
specialist leads on HumanEval+; at 17\,GB and above the 3-bit base is
smaller than the pruned 8-bit model and better on MBPP+.}
\label{fig:tradeoff}
\end{figure}}

The table locates a crossover rather than a winner\arxivonly{
(Figure~\ref{fig:tradeoff})}. Wherever the budget
admits a 3-bit-or-better quantization of the full model (here, 17~GB and
up), quantization is the safer choice: the 17.1~GB Q3 base matches every
pruned configuration on HumanEval+ and beats the 19.75~GB pruned Q8 on
MBPP+ by a significant 4.8 points while being 2.6~GB smaller. Quantization
costs this model essentially nothing down to 3 bits (the Q3 base even
edges the Q8 base on MBPP+, 0.775 versus 0.772, i.e.\ noise), so pruning
has nothing to offer at budgets that fit it. Below that boundary the
ordering flips: 2-bit quantization visibly damages the full model, and
the pruned specialist at a comfortable 4--5 bits is the significantly
better HumanEval+ model at equal size, while MBPP+ becomes a tie. We read
the residual benchmark split as informative about what pruning removes:
coding-calibrated pruning keeps the experts HumanEval's core generation
exercises, while MBPP's broader task spread leans on capability spread
across more experts, which quantization degrades gracefully and pruning
deletes. The guidance is simple: quantize down to 3 bits before pruning
anything; prune below that boundary, or when context headroom matters more
than the last MBPP+ points. The two axes also compose rather than compete:
every pruned point in Table~\ref{tab:iso} is itself quantized, and the
pruned Q5 model is the strongest sub-13.5~GB configuration we measured,
significantly ahead of the 2-bit full model on HumanEval+.

On throughput, the pruned Q4 model generates at 85 tokens/s and the Q8 at 61
tokens/s on our device, against 2{,}600 and 2{,}450 tokens/s of prompt
processing respectively. Pruning experts does not speed up generation here,
which is expected: on a bandwidth-bound unified-memory device, decode speed
tracks the active parameter count, and pruning changes the total, not the
per-token active set. What it buys is fit and the precision headroom of
Table~\ref{tab:iso}, not tokens per second.

\subsection{Stress-testing the crossover}
\label{sec:stress}

% Numbers: runs/strat/{s1_onpolicy,s3_guard,s5_ablation}/eval.json,
% runs/strat/s5_ablation/splithalf.json, runs/d0/attribution.json,
% runs/e1/eval.json, runs/e1/analysis.json. Bars: runs/qwen36/eval.json.
The crossover invites an obvious objection: perhaps we simply pruned
badly, and a better recipe would let the deep specialist beat the 2-bit
base. We attacked our own result with five follow-ups on Qwen3.6, each
with its kill criterion recorded in the project's run ledger before
launch. All five strengthen the crossover instead of overturning it.

\paragraph{Better calibration and guarded selection.} Re-profiling on the
base model's own greedy outputs (on-policy calibration, removing the
teacher-forcing mismatch) lifts the 25\%-keep specialist from 0.787 to
0.817 HumanEval+; protecting the top general-traffic experts before
filling the budget by saliency (a union guard) reaches 0.829. Both beat
the plain 25\% baseline, and both remain more than six HumanEval+ points
below the 2-bit base (0.896) while a 37.5\%-keep on-policy point (0.835)
trails it by six points using \emph{more} memory. Better recipes narrow
the gap to the pruning baseline, not to quantization.

\paragraph{Causally measured importance.} Since routing statistics are
proxies, we measured expert importance directly: 2{,}000 antithetic pairs
of random half-masks (each expert ablated at the router in exactly one run
of a pair, on identical tokens), importance as the paired loss difference.
The estimator came out noise-dominated at this sample size: split-half
reliability $r=0.05$ per layer, and the top-64 sets selected by the two
halves overlap at 16.7 of 64, exactly the 16.0 expected by chance; the
resulting checkpoint is degenerate. A causal signal at expert granularity
is too diffuse to extract at a compute budget far exceeding what any
practitioner would spend on selection, which points at the mechanism
below rather than at better scoring.

\paragraph{Failure attribution.} If deep pruning failed by deleting
specifically the experts some items need, the base model should route
failed items into the deleted set more than solved ones. It barely does:
teacher-forcing the base over its own solutions, items the 25\%-keep
model fails route 1.0 points more of their gate mass into deleted experts
than items it solves (95\% CI $[-0.7,+2.6]$, pooled; a small significant
+2.5 on MBPP+ only), against a baseline of 49--60\% of mass into
deleted experts on average. Deep pruning's damage is diffuse
capacity loss, not item-specific expert deletion, which is why better
selection cannot repair it.

\paragraph{Agentic evaluation.} Single-shot benchmarks might understate a
compressed model that could recover in a deployment loop with execution
feedback. We gave each model one repair turn per failed item, showing it
the failing test input and expected output, on the pooled 542-item
HumanEval+/MBPP+ set. Compression does repair
disproportionately: every compressed variant gains more from the repair
turn than the Q8 reference (paired bootstrap on the pooled
set: 25\%-keep $+3.9$ points more, CI $[+1.1,+6.6]$; 2-bit base $+2.8$,
CI $[+0.9,+4.8]$; the three compressed variants jointly $+2.9$, CI
$[+1.2,+4.6]$, $p=0.0005$), and the repair turn erases the 2-bit
quantization penalty outright: the 13\,GB 2-bit base ends at 0.860 pooled
pass@1 against 0.858 for the 38\,GB Q8 reference it trailed single-shot.
But the crossover ordering survives the agentic setting: the pruned
specialist's extra gain over the 2-bit base is not statistically
distinguishable from zero (CI $[-1.8,+4.2]$), and it still ends 6.5
points behind at less than its memory. Absolute repair rates are
inflated by test-visible feedback and identical prompting across models;
only the between-model comparison carries the claim. Single-shot evaluations overstate
compression penalties in general; they do not misorder pruning against
quantization.

\subsection{Recovery tuning}
\label{sec:heal}

The aggressive 25\%-keep specialist is the natural target for recovery
tuning: it carries the largest coding gap to the base model (10.3 points
HumanEval+), and it is small enough (10.9B parameters) to fine-tune on the
same single device it will be served from. We heal it by self-distillation:
the base model answers the 2{,}047 calibration prompts (none empty), and
the pruned model is trained on the 1{,}863 of those exchanges that fit the
training sequence length, with the loss masked to the assistant turn. One architectural constraint shapes the method and sharpens the
comparison to \citet{martin2026translation}: the routed experts are stored
as fused three-dimensional parameters that the LoRA implementation cannot
wrap, so we adapt every other linear projection (attention, the
linear-attention projections, the shared expert) with LoRA rank 64, train
the router matrices in full, and the routed experts stay \emph{exactly}
frozen (verified byte-identical after training). Recovery
thus comes entirely from re-routing and from the shared, always-on pathway,
never from editing the surviving experts.

This suffices, with the strongest evidence coming from perplexity and a
modest but consistent and significant signal from pass@1. Held-out code
perplexity falls from 5.45 to 4.85, below even the base model's 5.38 on this
set; the crossing itself is expected rather than remarkable, since the healed
model is the only one fine-tuned on base-generated responses drawn from the
same distribution as this holdout, and we read it as evidence the
distillation took, not as superiority to base. The collateral damage to
general text, which pruning had driven from 5.20 to 30.9, recovers most of
the way back to 9.4; these are large, unambiguous moves. Pass@1 improves on both code benchmarks we test:
HumanEval+ from 0.787 to 0.841 (recovering 52\% of the gap to base, net
$+9$ problems), and MBPP+ from 0.659 to 0.685 (recovering 23\% of its gap,
net $+10$ problems). Neither single-benchmark gain is decisive on its own
($n=164$ and $n=378$ respectively), but the effect is consistent in
direction, and pooled across both benchmarks the recovery is
significant: 43 problems flip to passing against 24 lost (exact McNemar
$p=0.027$), a $+3.5$-point mean improvement. We deliberately do not claim a
larger pass@1 effect than this; the perplexity recovery is the dominant
evidence and the pass@1 gains corroborate it. Trainable parameters are
95.2M, 0.86\% of the model, and healing runs in under three hours on the
target device.

This stands in pointed contrast to the translation-domain finding of
\citet{martin2026translation} (their \S4.3: ``parameter-efficient methods
are insufficient to recover entire pruned parameter blocks''), that full
fine-tuning is required. In the coding domain, with
the routed experts held frozen, parameter-efficient recovery restores the
majority of the lost capability. We read the difference as domain-dependent
MoE plasticity: what a coding specialist loses under aggressive pruning is
recoverable through routing and shared-pathway adaptation alone, without
touching the experts that pruning kept.

A methodological note that materially affected this result: we twice
observed a recovery experiment silently evaluating the \emph{unhealed}
model (a save format a downstream idempotency check re-pruned over, and a
precision path that rounded small updates away at merge time), with the
healed checkpoint scoring identically to the unhealed one to many decimal
places. A byte-level tensor diff plus a frozen-parameter check is
therefore a required gate before believing any recovery number, and we
report both here.

\section{Limitations and Future Work}

Our study covers two model families under one controlled protocol; the
criterion reversal of \secref{sec:gemma}{cross-family} is exactly the kind of
finding that argues for more, and we make no claim that a third family
would side with either of ours. Recovery tuning was run on one family.
Scaling further exposes a systems boundary worth naming. Our pipeline
profiles by loading the model with expert-usage hooks, which requires the
model to fit in memory. This is not an issue for the class of coding MoEs
that already fit the target hardware, but the largest open models
(hundreds of billions of parameters, distributed as block-quantized
checkpoints) exceed a single unified-memory device even before activations,
and standard offloaded loading stages the whole checkpoint before it can
spill to disk. Profiling models larger than memory therefore needs a
streaming, dequantize-per-shard profiler, which we leave to future work and
which we expect to also unlock pruning of the frontier-scale models where
memory pressure is most acute.

Two further limitations bound the present claims. Our calibration uses a
single seed and a single mix composition; the sensitivity of expert
selection to calibration randomness is quantified only by the
criterion-overlap analysis. The calibration mix includes Evol-CodeAlpaca,
an instruction corpus whose problems may overlap HumanEval-style
benchmarks. A normalized 13-gram scan finds shared n-grams with the
calibration mix for 9 of 164 HumanEval+ and 2 of 378 MBPP+ problems (8 and
1 against the healing distillation set). The pruning comparisons are
unaffected regardless (the base model is never trained), and the healing
result survives the check directly: of the 43 problems the healed model
newly passes across both benchmarks, exactly one lies in the overlap
set. \arxivonly{Uniform per-layer capacity is a
deployment-driven simplification; dynamic allocation helps in translation
\citep{martin2026translation} and may compose with our findings.}
General-text perplexity is a coarse proxy for what is lost; we do not
claim the specialists are safe general assistants, and
\citet{biomed2026reliability} suggest reliability audits before any
deployment beyond coding. All throughput numbers come from one device
class; relative sizes transfer, absolute speeds need not.

\section{Conclusion}

Open MoE models put the best coding assistants behind a memory wall
unrelated to the compute a coding session actually uses. On two model
families, one-shot expert pruning with coding-focused calibration removes
half the experts while leaving HumanEval+ statistically unchanged; on
MBPP+ the same point is free on one family and costs a significant 4.5
points on the other. Which experts to keep decides this, and no published
criterion decides it portably: the criterion that dominates Qwen loses by
a significant 12.8 points on Gemma to one that collapses on Qwen, so
one-shot pruning results, including ours, are family-specific until
functionally validated per model.
What does transfer: random selection collapses on both families, quality
falls as contrast displaces magnitude on both, and perplexity misleads on
both, up to a pruned model that beats base perplexity while losing 58
points of pass@1. At the most aggressive prune, parameter-efficient recovery with the
routed experts frozen recovers half the HumanEval+ gap, contrasting with
its reported failure for translation. Against quantization the answer is a crossover: quantize down to 3 bits
before pruning anything, and prune below that boundary; the open problem
is profiling models larger than memory.

\bibliographystyle{plainnat}
\bibliography{refs}

\end{document}